\documentclass{article} 
\usepackage{iclr2022_conference,times}

\usepackage{hyperref}
\usepackage{url}
\usepackage{paralist}
\usepackage{booktabs}

\title{1st ICLR International Workshop on Privacy, Accountability, Interpretability, Robustness, Reasoning on Structured Data ($\mathbf{PAIR}^2$Struct)}

\iclrfinalcopy

\author{Hao Wang\textsuperscript{1}, Wanyu Lin\textsuperscript{2}, Hao He\textsuperscript{3}, Di Wang\textsuperscript{4}, Chengzhi Mao\textsuperscript{5}, Muhan Zhang\textsuperscript{6} \\
\textsuperscript{1}Rutgers University, \textsuperscript{2}Hong Kong Polytechnic University, \textsuperscript{3}Massachusetts Institute of Technology\\
\textsuperscript{4}King Abdullah University of Science and Technology, \textsuperscript{5}Columbia University, \textsuperscript{6}Peking University
\\
\texttt{hw488@cs.rutgers.edu}, \texttt{ wan-yu.lin@polyu.edu.hk}, \texttt{ haohe@mit.edu},\\
\texttt{di.wang@kaust.edu.sa}, \texttt{mcz@cs.columbia.edu}, \texttt{muhan@wustl.edu}\\
}

\begin{document}

\maketitle

\begin{abstract}
Recent years have seen advances on principles and guidance relating to accountable and ethical use of artificial intelligence (AI) spring up around the globe. Specifically, Data {\bf P}rivacy, {\bf A}ccountability, {\bf I}nterpretability, {\bf R}obustness, and {\bf R}easoning have been broadly recognized as fundamental principles of using machine learning (ML) technologies on decision-critical and/or privacy-sensitive applications. On the other hand, in tremendous real-world applications, data itself can be well represented as various structured formalisms, such as graph-structured data (e.g., networks), grid-structured data (e.g., images), sequential data (e.g., text), etc. By exploiting the inherently structured knowledge, one can design plausible approaches to identify and use more relevant variables to make reliable decisions, thereby facilitating real-world deployments.  
\end{abstract}
\section{Introduction} 

In this workshop, we will examine the research progress towards accountable and ethical use of AI from diverse research communities, such as the ML community, security \& privacy community, and more. Specifically, we will focus on the limitations of existing notions on {\bf P}rivacy, {\bf A}ccountability, {\bf I}nterpretability, {\bf R}obustness, and {\bf R}easoning. We aim to bring together researchers from various areas (e.g.,  ML, security \& privacy, computer vision, and healthcare) to facilitate discussions including related challenges, definitions, formalisms, and evaluation protocols regarding the accountable and ethical use of ML technologies in high-stake applications with structured data. In particular, we will discuss the {\bf interplay} among the fundamental principles from theory to applications. We aim to identify new areas that call for additional research efforts. Additionally, we will seek possible solutions and associated interpretations from the notion of causation, which is an inherent property of systems. We hope that the proposed workshop is fruitful in building accountable and ethical use of AI systems in practice. 

The following is a non-exhaustive list of topics we aim to address through our invited talks, panels, and accepted papers:

\begin{itemize}
    \item Privacy-preserving machine learning methods on structured data (e.g., graphs, manifolds, images, and text). 

    \item Theoretical foundations for privacy-preserving and/or explainability of deep learning on structured data (e.g., graphs, manifolds, images, and text).  

    \item Interpretability and accountability in different application domains including healthcare, bioinformatics, finance, physics, etc.  

    \item Improving interpretability and accountability of black-box deep learning with graphical abstraction (e.g., causal graphs, graphical models, computational graphs). 
    
    \item Robust machine learning methods via graphical abstraction (e.g., causal graphs, graphical models, computational graphs).

    \item Relational/graph learning under robustness constraints (robustness in face of adversarial attacks, distribution shift, environment changes, etc.).   

    \item Relational or object-oriented reasoning for computer vision, natural language processing, robotics, data mining, etc. 

    \item New benchmark datasets and evaluation metrics.  
\end{itemize}


\section{Tentative Schedule}
We plan to include 7 invited talks (25 minutes + 5 minutes questions) and 8 contributed talks by authors of submitted works. To further encourage the
exchange of ideas and discussions, the workshop will conclude with a panel including all of our
invited speakers and other distinguished panelists. We will collect questions via an online form
from the audience and online (streaming) viewers. 

\begin{table}[h]
\center
\begin{tabular}{l|l}
\toprule
9:00 - 9:15 & Introduction and Opening Remarks \\ \midrule
9:15 - 9:45 & Invited Talk 1                   \\ \midrule
9:45 - 10:15 & Invited Talk 2                  \\ \midrule
10:15 -10:20& Contributed Talk 1               \\ \midrule
10:20 -10:25& Contributed Talk 2               \\ \midrule
10:25 -11:00 & Invited Talk 3                 \\ \midrule
11:00 -11:30& Invited Talk 4                \\ \midrule
11:30 -11:35& Contributed Talk 3               \\ \midrule
11:35 -11:40& Contributed Talk 4               \\ \midrule
11:40 -14:00& Poster Session 1                    \\ \midrule
14:00 -14:30& Invited Talk 5                   \\ \midrule
14:30 -15:00& Invited Talk 6                   \\ \midrule
15:00 -15:05& Contributed Talk 5               \\ \midrule
15:05 -15:10& Contributed Talk 6                \\ \midrule
15:10 -15:40& Invited Talk 7                     \\ \midrule
15:40 -15:45& Contributed Talk 7                 \\ \midrule
15:45 -15:50& Contributed Talk 8               \\ \midrule
15:50 -16:30& Panel Discussion                 \\ \midrule
16:30 -18:00& Poster Session 2                  \\ \bottomrule
\end{tabular}
\end{table}


\section{Organizers and Speakers}


\textbf{Hao Wang} is an Assistant Professor in the Department of Computer Science at Rutgers University. Previously he was a Postdoctoral Associate at the Computer Science and Artificial Intelligence Lab (CSAIL) of MIT, working with Dina Katabi and Tommi Jaakkola. He has been a visiting researcher in the Machine Learning Department of Carnegie Mellon University. His research focuses on statistical machine learning, deep learning, and data mining, with broad applications on recommender systems, healthcare, user profiling, social network analysis, text mining, etc. He was the recipient of the Microsoft Fellowship in Asia, the Baidu Research Fellowship, and Amazon Faculty Research Award. He is the co-organizer of the ICML 2019 Workshop on ``Learning and Reasoning with Graph-Structured Representations'' and the CVPR 2019 Workshop on ``Towards Causal, Explainable and Universal Medical Visual Diagnosis''.   

\textbf{Wanyu Lin} is a Research Assistant Professor in the Department of Computing at The Hong Kong Polytechnic University. Wanyu’s main research interest is in developing machine learning algorithms for graph-structured data. Currently, her research activities focus on model explanations for graph neural networks and trustworthy machine learning on graphs. Wanyu’s methods of graph representation learning have been applied to various applications, such as computer vision and social networks. 

\textbf{Hao He} is a PhD candidate at Massachusetts Institute of Technology, advised by Prof. Dina Katabi. 
He received a B.Sc. in EECS from Peking University. His research focuses on machine learning algorithms that are generalizable across domains, robust to security threats, fair to sub-populations with real world applications in healthcare and smart home. Hao serves as a reviewer for ICML, NeurIPS, ICLR, CVPR, ICCV, AAAI, UAI, IJCAI.  

\textbf{Di Wang} is an Assistant Professor at King Abdullah University of Science and Technology. His research focuses on privacy-preserving machine learning, robust statistics, interpretable machine learning, optimization and high dimensional statistics. He authored several publications at major machine learning conferences/journals such as ICML, NeurIPS, JMLR, ALT,  TIT, AAAI, IJCAI. And he serves as a reviewer/program committee for JASA, STOC, TDSC, TIFS, IEEE S\&P, ICML, NeurIPS, ICLR, AISTATS, CVPR, ECCV, ICCV, AAAI, IJCAI.  

\textbf{Chengzhi Mao} is a fourth-year PhD student at Columbia University advised by Prof. Carl Vondrick and Prof. Junfeng Yang. His research focuses on the Robustness, Interpretability, and Causality, of computer vision models. He first-authored several publications at major machine learning and computer vision conferences, such as NeurIPS, CVPR, ICCV, and ECCV. He serves as a reviewer for ICML, NeurIPS, CVPR, ICCV, AAAI, ECCV, and ICLR.  

\textbf{Muhan Zhang} is an Assistant Professor at Peking University. His research interests lie in machine learning over graphs, in particular the algorithms, theories and applications of graph neural networks. He has published multiple pioneering works in the field of graph neural networks on top conferences such as NeurIPS, ICLR, AAAI, KDD. He serves as a reviewer regularly for NeurIPS, ICML, ICLR, AAAI, IJCAI, TPAMI, TNNLS, TKDE, AOAS, JAIR, etc.

\subsection{Invited Speakers}

\textbf{Bo Li} is an Assistant Professor in the Computer Science Department at University of Illinois at Urbana-Champaign. Her research focuses on machine learning, {\em security}, {\em privacy}, and game theory. She is particularly interested in exploring vulnerabilities of machine learning systems to various adversarial attacks and endeavors to develop real-world {\em robust learning systems}. She has explored different types of adversarial attacks, including evasion and poisoning attacks in digital and physical worlds with different constraints. She has developed robust learning algorithms based on game theory, prior knowledge of data distribution, and properties of learning tasks. Her work directly benefits applications such as computer vision, natural language processing, audio recognition, and privacy-preserving machine learning systems.  

\textbf{Reza Shokri} is an Assistant Professor in the Computer Science Department at National University of Singapore. His research focuses on data privacy and trustworthy machine learning. He is interested in designing methods to quantitatively measure the privacy risks of data processing algorithms, and build scalable schemes for generalizable machine learning models that are also {\em privacy-preserving}, {\em robust}, {\em interpretable}, and fair. His research is on analyzing the trade-offs between different pillars of trust in machine learning for practical scenarios, and on resolving such conflicts with rigorous mathematical guarantees. He has been working on many interesting problems in this domain, including {\em trustworthy} federated learning, differential privacy for machine learning, fairness versus privacy in machine learning, privacy-aware model explanations, privacy-preserving data synthesis, and quantifying privacy risks of data analytics.

\textbf{Elias Bareinboim} is an Associate Professor in the Department of Computer Science and the director of the Causal Artificial Intelligence (CausalAI) Laboratory at Columbia University. His research focuses on causal and counterfactual inference and their applications to data-driven fields in the health and social sciences as well as artificial intelligence and machine learning. His work was the first to propose a general solution to the problem of ``data-fusion,'' providing practical methods for combining datasets generated under different experimental conditions and plagued with various biases. More recently, Bareinboim has been exploring the intersection of {\em causal inference with decision-making} (including reinforcement learning) and {\em explainability} (including fairness analysis). Before joining Columbia, he was an assistant professor at Purdue University and received his Ph.D. in Computer Science from the University of California, Los Angeles. Bareinboim was named one of ``AI's 10 to Watch'' by IEEE, and is a recipient of an NSF CAREER Award, the Dan David Prize Scholarship, the 2014 AAAI Outstanding Paper Award, and the 2019 UAI Best Paper Award.

%

\textbf{Yang Zhang} is a faculty member at CISPA Helmholtz Center for Information Security. Previously, he was a research group leader at CISPA. His research focuses on the security and {\em privacy} of machine learning, social network analysis, and algorithmic fairness. He is particularly interested in quantifying and mitigating the privacy risks of machine learning models. He has been working on many interesting problems, including privacy/property inference attacks to various machine learning models (e.g., Generative Adversarial Networks, Graph Neural Networks),  data synthesis with differential privacy, etc.

\textbf{Jiajun Wu} is an Assistant Professor of Computer Science at Stanford University, working on computer vision, machine learning, and computational cognitive science. Before joining Stanford, he was a Visiting Faculty Researcher at Google Research. He received his Ph.D. in Electrical Engineering and Computer Science at Massachusetts Institute of Technology. He is particularly interested in {\em machine perception}, {\em reasoning}, and interaction with the physical world, drawing inspiration from human cognition. His research has been recognized through the ACM Doctoral Dissertation Award Honorable Mention, the AAAI/ACM SIGAI Doctoral Dissertation Award, the MIT George M. Sprowls Ph.D. Thesis Award in Artificial Intelligence and Decision-Making, the 2020 Samsung AI Researcher of the Year, the IROS Best Paper Award on Cognitive Robotics, and faculty research awards and graduate fellowships from Samsung, Amazon, Facebook, Nvidia, and Adobe. 

\textbf{Lei Xing} is currently the Jacob Haimson Professor of Medical Physics and Director of Medical Physics Division of Radiation Oncology Department at Stanford University. He also holds affiliate faculty positions in Departments of Electric Engineering (courtesy), Biomedical Informatics, Bio-X, and Molecular Imaging Program at Stanford. His research has been focused on inverse treatment planning, artificial intelligence (AI) in {\em medicine, tomographic image reconstruction, CT, optical and PET imaging instrumentations, image-guided interventions, nanomedicine, and applications of molecular imaging in radiation oncology}. Dr. Xing is an author on more than 300 peer-reviewed publications, a co-inventor on many issued and pending patents, and a co-investigator or principal investigator on numerous NIH, DOD, NSF, ACS grants, and projects from other funding agencies and corporates. He and his lab members have received numerous awards from ACS, AAPM, ASTRO, WMIC, RSNA, and industrial companies such as Varian, Google, NVIDIA, and Siemens in the past decade. 
He is a fellow of AAPM and AIMBE (American Institute for Medical and Biological Engineering).   

\textbf{Zachary Chase Lipton} is an Assistant Professor of Operations Research and Machine Learning at Carnegie Mellon University. His research spans core machine learning methods and their social impact and addresses diverse application areas, including {\em clinical medicine} and natural language processing. Current research focuses include {\em robustness} under distribution shift, breast cancer screening, the effective and equitable allocation of organs, and the intersection of {\em causal thinking} with messy data. He is the founder of the Approximately Correct (\url{approximatelycorrect.com}) blog and the creator of Dive Into Deep Learning, an interactive open-source book drafted entirely through Jupyter notebooks.

\section{Previous Related Workshops}
\vspace{-6pt}

This workshop provides a platform to foster collaborations and a thorough understanding of responsible and trustworthy AI from the perspectives of privacy, accountability, interpretability, robustness, and reasoning, with a focus on structured data (as opposed to low-dimensional tabular data). It is closely related to previous workshops such as the \emph{Uncertainty and Robustness in Deep Learning} (ICML 2021), \emph{Interpretable Machine Learning in Healthcare} (ICML 2021), \emph{Responsible AI} (ICLR 2021), \emph{XXAI: Extending Explainable AI Beyond Deep Models and Classifiers} (ICML 2020), \emph{Interpretable Inductive Biases and Physically Structured Learning} (NeurIPS 2020), and \emph{Robust AI in Financial Services: Data, Fairness, Explainability, Trustworthiness, and Privacy} (NeurIPS 2019). Most previous workshops either primarily focus on one single aspect of responsible AI (such as interpretability or robustness) or lack emphasis on the intrinsic structures (such as graphs, images, and natural languages) of data. To fill in the blanks, this workshop advocates the importance of (1) considering the interplay among different aspects of responsible AI in a holistic way, such that optimal trade-off among these aspects can be achieved, and (2) developing structure-aware responsible AI, such that attack on privacy leveraging the intrinsic structure in data can be avoided and accountability/interpretability/robustness/reasoning can be optimized by leveraging structures in data. Therefore in contrast to previous workshops, this workshop is uniquely positioned as a convergent venue where researchers studying different aspects of responsible AI (for data with different structures) can exchange ideas, initiate collaborations, and together push the limit of responsible and trustworthy AI. 


\section{Diversity Commitment}
We actively encouraged all forms of diversity when selecting organizers and inviting speakers. The final groups of organizers and speakers are diverse concerning gender, race, affiliations, nationality and scientific background. The two groups cover a full scale of scientific seniority, from PhD students to assistant and full professors.

\section{Access}

Due to COVID-19, we will livestream all talks and discussion sessions. We will publish the accepted papers and the slides of the speakers on the workshop website. We will include a bibliography of research papers most relevant to the workshop so that the attendants can quickly touch both the classic works and the frontiers of the field. We will also include the links to previous related workshops. To increase our impact, we plan to post the information of our workshop to online media such as Twitter, Facebook, Reddit, Weibo, and Zhihu. To fund these activities, we will solicit sponsorship from companies with data reliability concerns such as Facebook, Google, Amazon, and Salesforce. 



\section{Anticipated Audience Size and Plan to Get Audience}
We expect 100$\sim$200 participants and around $50$ workshop paper submissions. Advertisement as well as the link to the workshop official website will be sent out before and during the workshop via mailing lists in different institutions, Twitter, Facebook, etc. The diversity of the organizers ensures that the workshop information will reach out to potential participants from different institutions, different geographic locations, and different research areas.



\end{document}